\documentclass[a4paper,runningheads]{llncs}
\usepackage[latin9]{inputenc}
\usepackage{url}
\usepackage{amsmath}
\usepackage{amssymb}
\usepackage{graphicx}

\makeatletter

\providecommand{\tabularnewline}{\\}

\usepackage{graphicx}

\usepackage{url}
\urldef{\mailsa}\path{sotirios.chatzis@cut.ac.cy} 
\urldef{\mailsb}\path{c.partaourides@cut.ac.cy}

\@ifundefined{showcaptionsetup}{}{%
 \PassOptionsToPackage{caption=false}{subfig}}
\usepackage{subfig}
\makeatother

\begin{document}

\title{Deep Network Regularization via Bayesian Inference of Synaptic Connectivity}

\titlerunning{Deep Net Regularization via Bayesian Inference of Synaptic Connectivity}

\author{Harris Partaourides and Sotirios P. Chatzis}

\authorrunning{Harris Partaourides and Sotirios P. Chatzis}

\institute{Department of Electrical Eng., Computer Eng., and Informatics\\
 Cyprus University of Technology\\
\mailsb ; \mailsa}

\toctitle{Lecture Notes in Computer Science}

\tocauthor{Harris Partaourides and Sotirios P. Chatzis}
\maketitle
\begin{abstract}
Deep neural networks (DNNs) often require good regularizers to generalize
well. Currently, state-of-the-art DNN regularization techniques consist
in randomly dropping units and/or connections on each iteration of
the training algorithm. Dropout and DropConnect are characteristic
examples of such regularizers, that are widely popular among practitioners.
However, a drawback of such approaches consists in the fact that their
postulated probability of random unit/connection omission is a constant
that must be heuristically selected based on the obtained performance
in some validation set. To alleviate this burden, in this paper we
regard the DNN regularization problem from a Bayesian inference perspective:
We impose a sparsity-inducing prior over the network synaptic weights,
where the sparsity is induced by a set of Bernoulli-distributed binary
variables with Beta (hyper-)priors over their prior parameters. This
way, we eventually allow for marginalizing over the DNN synaptic connectivity
for output generation, thus giving rise to an effective, heuristics-free,
network regularization scheme. We perform Bayesian inference for the
resulting hierarchical model by means of an efficient Black-Box Variational
inference scheme. We exhibit the advantages of our method over existing
approaches by conducting an extensive experimental evaluation using
benchmark datasets.
\end{abstract}

\section{Introduction}

In the last few years, the field of machine learning has experienced
a new wave of innovation; this is due to the rise of a family of modeling
techniques commonly referred to as deep neural networks (DNNs) \cite{lecun}.
DNNs constitute large-scale neural networks, that have successfully
shown their great learning capacity in the context of diverse application
areas. Since DNNs comprise a huge number of trainable parameters,
it is key that appropriate techniques be employed to prevent them
from overfitting. Indeed, it is now widely understood that one of
the reasons behind the explosive success and popularity of DNNs
consists in the availability of simple, effective, and efficient regularization
techniques, developed in the last few years \cite{lecun}. 

Dropout is a popular regularization technique for (dense-layer) DNNs \cite{dropout}. In
essence, it consists in randomly dropping different units of the network
on each iteration of the training algorithm. This way, only the parameters
related to a subset of the network units are trained on each iteration;
this ameliorates the associated network overfitting tendency, and
it does so in a way that ensures that all network parameters are effectively
trained. In a different vein, \cite{dropconnect} proposed randomly
dropping DNN synaptic connections, instead of network units (and all
the associated parameters); they dub this approach DropConnect. As
showed therein, such a regularization scheme yields better results
than Dropout in several benchmark datasets, while offering provable
bounds of computational complexity.

Despite these merits, one drawback of these regularization schemes
can be traced to their very foundation and rationale: The postulated
probability of random unit/connection omission (e.g., dropout rate)
is a constant that must be heuristically selected; this is effected
by evaluating the network's predictive performance under different
selections of this probability, in some validation set, and retaining
the best performing value. This drawback has recently motivated research
on the theoretical properties of these techniques. Indeed, recent
theoretical work at the intersection of deep learning and Bayesian
statistics has shown that Dropout can be viewed as a simplified approximate
Bayesian inference algorithm, and enjoys links with Gaussian process
models under certain simplistic assumptions (e.g., \cite{understandDropout,bayes-dropout}). 

These recent results form the main motivation behind this paper. Specifically,
the main question this work aims to address is the following: Can
we devise an effective DNN regularization scheme, that marginalizes
over all possible configurations of network synaptic connectivity
(i.e., active synaptic connections), with the posterior over them
being inferred from the data? To address this problem, in this paper,
for the first time in the literature, we regard the DNN regularization
problem from the following Bayesian inference perspective: We impose
a sparsity-inducing prior over the network synaptic weights, where
the sparsity is induced by a set of Bernoulli-distributed binary variables.
Further, the parameters of the postulated Bernoulli-distributed binary
variables are imposed appropriate Beta (hyper-)priors, which give
rise to a full hierarchical Bayesian treatment for the proposed model. 

Under this hierarchical Bayesian construction, we can derive appropriate
posteriors over the postulated binary variables, which essentially
function as indicators of whether some (possible) synaptic connection
is retained or dropped from the network. Once these posteriors are
obtained using some available training data, prediction can be performed
by averaging (under a Bayesian inference sense) over multiple (posterior)
samples of the network configuration. This inferential setup constitutes
the main point of differentiation between our approach and DropConnect.
For simplicity, and to facilitate reference, we dub our approach DropConnect++.
We derive an efficient inference algorithm for our model by resorting
to the Black-Box Variational Inference (BBVI) scheme \cite{bbvi}.

The remainder of this paper is organized as follows: In Section 2,
we provide a brief overview of the theoretical background of our approach.
Specifically, we first briefly review DropConnect, which is the existing
work closest related to our approach; subsequently, we review the
inferential framework that will be used in the context of the proposed
approach, namely BBVI. In Section 3, we introduce our approach, and
derive its inference and prediction generation algorithms. Next, we
perform an extensive experimental evaluation of our approach, and
compare to popular (dense-layer) DNN regularization approaches, including
Dropout and DropConnect. To this end, we consider a number of well-established
benchmarks in the related literature. Finally, in the concluding section,
we summarize our contribution and discuss our results.

\section{Theoretical Background}

\subsection{DropConnect}

As discussed in the Introduction, DropConnect is a generalization
of Dropout under which each connection, rather than each unit, may
be dropped with some heuristically selected probability. Hence, the
rationale of DropConnect is similar to that of Dropout, since both
introduce dynamic sparsity within the model. Their core difference
consists in the fact that Dropout imposes sparsity on the output vectors
of a (dense) layer, while DropConnect imposes sparsity on the synaptic
weights $\boldsymbol{W}$. 

Note that this is not equivalent to setting $\boldsymbol{W}$ to be
a fixed sparse matrix during training. Indeed, for a DropConnect layer,
the output is given as \cite{dropconnect}: 
\begin{equation}
\boldsymbol{r}=a((\boldsymbol{Z}\circ\boldsymbol{W})\boldsymbol{v})
\end{equation}
where $\circ$ is the elementwise product, $a(\cdot)$ is the adopted
activation function, $\boldsymbol{W}$ is the matrix of synaptic weights,
$\boldsymbol{v}$ is the layer input vector, and $\boldsymbol{r}$
is the layer output vector. Further, $\boldsymbol{Z}$ is a matrix
of binary variables (indicators) encoding the connection information,
with
\begin{equation}
[\boldsymbol{Z}]_{i,j}\sim\mathrm{Bernoulli}(p)
\end{equation}
where $p$ is a heuristically selected probability. Hence, DropConnect
is a generalization of Dropout to the full connection structure of
a layer \cite{dropconnect}.

Training of a DropConnect layer begins by selecting an example $\boldsymbol{v}$,
and drawing a mask matrix $\boldsymbol{Z}$ from a $\mathrm{Bernoulli}(p)$
distribution to mask out elements of both the weight matrix and the
biases in the DropConnect layer. The parameters throughout the model
can be updated via stochastic gradient descent (SGD), or some modern
variant of it, by backpropagating gradients of the postulated loss
function with respect to the parameters. To update the weight matrix
$\boldsymbol{W}$ in a DropConnect layer, the mask is applied to the
gradient to update only those elements that were active in the forward
pass. Additionally, when passing gradients down, the masked weight
matrix $\boldsymbol{Z}\circ\boldsymbol{W}$ is used. 

\subsection{BBVI}

In general, Bayesian inference for a statistical model can be performed
either exactly, by means of Markov Chain Monte Carlo (MCMC), or via
approximate techniques. Variational inference is the most widely used
approximate technique; it approximates the posterior with a simpler
distribution, and fits that distribution so as to have minimum Kullback-Leibler
(KL) divergence from the exact posterior \cite{jaakola}. This way,
variational inference effectively converts the problem of approximating
the posterior into an optimization problem.

One of the significant drawbacks of traditional variational inference
consists in the fact that its objective entails posterior expectations
which are tractable only in the case of conjugate postulated models.
Hence, recent innovations in variational inference have attempted
to allow for rendering it feasible even in cases of more complex,
non-conjugate model formulations. Indeed, recently proposed solutions
to this problem consist in using stochastic optimization, by forming
noisy gradients with Monte Carlo (MC) approximation. In this context,
a number of different techniques have been proposed so as to successfully
reduce the unacceptably high variance of conventional MC estimators.
BBVI \cite{bbvi} is one of these recently proposed alternatives,
amenable to non-conjugate probabilistic models that entail both discrete
and continuous latent variables. 

Let us consider a probabilistic model $p(\boldsymbol{x},\boldsymbol{z})$
with observations $\boldsymbol{x}$ and latent variables $\boldsymbol{z}$,
as well as a sought variational family $q(\boldsymbol{z};\boldsymbol{\phi})$.
BBVI optimizes an evidence lower bound (ELBO), with expression 
\begin{equation}
\mathrm{log}\,p(\boldsymbol{x})\geq\mathcal{L}(\boldsymbol{\phi})=\mathbb{E}_{q(\boldsymbol{z};\boldsymbol{\phi})}[\mathrm{log}\,p(\boldsymbol{x},\boldsymbol{z})-\mathrm{log}\,q(\boldsymbol{z};\boldsymbol{\phi})]
\end{equation}
This is performed by relying on the ``log-derivative trick\textquotedblright{}
\cite{glynn,reinforce} to obtain MC estimates of the gradient. Specifically,
by application of the identities 
\begin{equation}
\nabla_{\boldsymbol{\phi}}q(\boldsymbol{z};\boldsymbol{\phi})=q(\boldsymbol{z};\boldsymbol{\phi})\nabla_{\boldsymbol{\phi}}\mathrm{log}\,q(\boldsymbol{z};\boldsymbol{\phi})
\end{equation}
\begin{equation}
\mathbb{E}_{q(\boldsymbol{z};\boldsymbol{\phi})}[\nabla_{\boldsymbol{\phi}}\mathrm{log}\,q(\boldsymbol{z};\boldsymbol{\phi})]=0
\end{equation}
the gradient of the ELBO (3) reads 
\begin{equation}
\nabla_{\boldsymbol{\phi}}\mathcal{L}(\boldsymbol{\phi})=\mathbb{E}_{q(\boldsymbol{z};\boldsymbol{\phi})}[f(\boldsymbol{z})]
\end{equation}
where 
\begin{equation}
f(\boldsymbol{z})=\nabla_{\boldsymbol{\phi}}\mathrm{log}\,q(\boldsymbol{z};\boldsymbol{\phi})\left[\mathrm{log}\,p(\boldsymbol{x},\boldsymbol{z})-\mathrm{log}\,q(\boldsymbol{z};\boldsymbol{\phi})\right]
\end{equation}

The so-obtained MC estimator, based on computing the posterior expectations
$\mathbb{E}_{q(\boldsymbol{z};\boldsymbol{\phi})}[\cdot]$ via sampling
from $q(\boldsymbol{z};\boldsymbol{\phi})$, only requires evaluating
the log-joint distribution $\mathrm{log}\,p(\boldsymbol{x},\boldsymbol{z})$,
the log-variational distribution $\mathrm{log}\,q(\boldsymbol{z};\boldsymbol{\phi})$,
and the score function $\nabla_{\boldsymbol{\phi}}\mathrm{log}\,q(\boldsymbol{z};\boldsymbol{\phi})$,
which is easy for a large class of models. However, the resulting
estimator may have high variance, especially if the variational approximation
$q(\boldsymbol{z};\boldsymbol{\phi})$ is a poor fit to the actual
posterior. In order to reduce the variance of the estimator, one common
strategy in BBVI consists in the use of \emph{control variates. }

A control variate is a random variable that is included in the estimator,
preserving its expectation but reducing its variance. The most usual
choice for control variates, which we adopt in this work, is the so-called
weighted score function: Under this selection, the ELBO gradient becomes
\begin{equation}
\nabla_{\boldsymbol{\phi}}\mathcal{L}(\boldsymbol{\phi})=\mathbb{E}_{q(\boldsymbol{z};\boldsymbol{\phi})}[f(\boldsymbol{z})-\varpi h(\boldsymbol{z})]
\end{equation}
where the score function reads
\begin{equation}
h(\boldsymbol{z})=\nabla_{\boldsymbol{\phi}}\mathrm{log}\,q(\boldsymbol{z};\boldsymbol{\phi})
\end{equation}
while the weights $\varpi$ yield the (optimized) expression \cite{bbvi}
\begin{equation}
\varpi=\frac{\mathrm{Cov}\left(f(\boldsymbol{z}),h(\boldsymbol{z})\right)}{\mathrm{Var}\left(h(\boldsymbol{z})\right)}
\end{equation}
On this basis, derivation of the sought variational posteriors is
performed by utilizing the gradient expression (8) in the context
of popular, off-the-shelf optimization algorithms, e.g. AdaM \cite{adam}
and Adagrad \cite{adagrad}. 

\section{Proposed Approach}

The output expression of a DropConnect++ layer is fundamentally similar
to conventional DropConnect, and is given by (1). However, DropConnect++
introduces an additional hierarchical set of assumptions regarding
the matrix of binary (mask) variables $\boldsymbol{Z}=[z_{ij}]_{i,j}$,
which indicate whether a synaptic connection is inferred to be on
or off. 

Specifically, as usual in hierarchical graphical models, we assume
that the random matrix $\boldsymbol{Z}$ is drawn from an appropriate
prior; we postulate

\begin{equation}
p(\boldsymbol{Z}|\boldsymbol{\Pi})=\prod_{i,j}p(z_{ij}|\pi_{ij})=\prod_{i,j}\mathrm{Bernoulli}(z_{ij}|\pi_{ij})
\end{equation}
Subsequently, to \emph{facilitate further regularization} for
DropConnect++ layers under a Bayesian inferential perspective, 
the prior parameters $\pi_{ij}\triangleq p(z_{ij}=1)$
are imposed their own (hyper-)prior. Specifically, we elect to impose
a Beta hyper-prior, yielding
\begin{equation}
p(\pi_{ij}|\alpha,\beta)=\mathrm{Beta}(\pi_{ij}|\alpha,\beta),\;\forall i,j
\end{equation}
Under this definition, to train a postulated DNN incorporating DropConnect++
layers, we need to resort to some sort of Bayesian inference technique.
In this paper, we resort to BBVI, as we explain next.

\subsection{Training DNNs with DropConnect++ layers}

Let us consider a DNN the observed training data of which constitute
the set $\mathcal{D}=\{\boldsymbol{d}_{n}\}_{n=1}^{N}$. In case of
a generative modeling scheme, each example $\boldsymbol{d}_{n}$ is
a single observation, say $\boldsymbol{x}_{n}$, from the distribution
we wish to model. On the other hand, in case of a discriminative modeling
task, each example $\boldsymbol{d}_{n}$ is an input/output pair,
for instance $\boldsymbol{d}_{n}=(\boldsymbol{x}_{n},\boldsymbol{y}_{n})$.
In both cases, conventional DNN training consists in optimizing a
negative loss function, measuring the fit of the model to the training
dataset $\mathcal{D}$. Such measures can be equivalently expressed
in terms of a log-likelihood function $\mathrm{log}\,p(\mathcal{D})$;
under this regard, DNN training effectively boils down to maximum-likelihood
estimation \cite{dae,glorot}.

The deviation of a DNN comprising DropConnect++ layers from this simple
training scheme stems from obtaining appropriate \emph{posterior }distributions
over the \emph{latent} \emph{variables} of DropConnect++, namely the
binary indicator matrices of synaptic connectivity, $\boldsymbol{Z}$,
as well as the associated parameters with hyper-priors imposed over
them, namely the matrices of (prior) parameters $\boldsymbol{\Pi}$.
To this end, DropConnect++ postulates \emph{separate} posteriors over
each entry of the random matrices $\boldsymbol{Z}$, that correspond
to each individual synapse, $(i,j)$:
\begin{equation}
q(\boldsymbol{Z})=\prod_{i,j}q(z_{ij}|\tilde{\pi}_{ij}),\mathrm{\;with:}\qquad q(z_{ij}|\tilde{\pi}_{ij})=\mathrm{Bernoulli}(z_{ij}|\tilde{\pi}_{ij})
\end{equation}
 Further, we consider that the matrices of prior parameters, $\boldsymbol{\Pi}$,
yield a factorized (hyper-)posterior with Beta-distributed factors
of the form
\begin{equation}
q(\pi_{ij})=\mathrm{Beta}(\pi_{ij}|\tilde{\alpha}_{ij},\tilde{\beta}_{ij})
\end{equation}

Our construction entails a conditional log-likelihood term, $\mathrm{log}\,p(\mathcal{D}|\boldsymbol{Z})$.
This is similar to a conventional DNN, with the weight matrices $\boldsymbol{W}$
at each layer multiplied with the corresponding latent indicator (mask)
matrices, $\boldsymbol{Z}$ (in analogy to DropConnect). The corresponding
posterior expectation term, $\mathbb{E}_{q(\boldsymbol{Z})}[\mathrm{log}\,p(\mathcal{D}|\boldsymbol{Z})]$,
constitutes part of the ELBO expression of our model. Unfortunately,
this term is analytically intractable due to the entailed nonlinear
dependencies on the indicator matrix $\boldsymbol{Z}$, which stem
from the nonlinear activation function $a(\cdot)$. Following the
previous discussion, we ameliorate this issue by resorting to an efficient
approximation obtained by drawing MC samples. The so-obtained ELBO
functional expression eventually becomes:
\begin{equation}
\begin{aligned}\mathcal{L}(\mathcal{D})\approx & -\sum_{i,j}\mathrm{KL}\big[q(z_{ij}|\tilde{\pi}_{ij})||p(z_{ij}|\pi_{ij})\big]-\sum_{i,j}\mathrm{KL}\big[q(\pi_{ij}|\tilde{\alpha}_{ij},\tilde{\beta}_{ij})||p(\pi_{ij}|\alpha,\beta)\big]\\
 & +\frac{1}{L}\sum_{l,n=1}^{L,N}\mathrm{log}\,p(\boldsymbol{d}_{n}|\boldsymbol{Z}^{(l)})
\end{aligned}
\end{equation}
where $L$ is the number of samples, $\boldsymbol{Z}^{(l)}=[z_{ij}^{(l)}]_{i,j}$
and $z_{ij}^{(l)}\sim\mathrm{Bernoulli}(z_{ij}|\tilde{\pi}_{ij})$. 

This concludes the formulation of the proposed inferential setup for
a DNN that contains DropConnect++ layers. On this basis, inference
is performed by resorting to BBVI, which proceeds as described previously.
Denoting $\tilde{\boldsymbol{\pi}}=(\tilde{\pi}_{ij})_{i,j},\tilde{\boldsymbol{\alpha}}=(\tilde{\alpha}_{ij})_{i,j},\tilde{\boldsymbol{\beta}}=(\tilde{\beta}_{ij})_{i,j}$,
the used ELBO gradient reads
\begin{equation}
\begin{aligned}\nabla_{\tilde{\boldsymbol{\pi}},\tilde{\boldsymbol{\alpha}},\tilde{\boldsymbol{\beta}},\boldsymbol{W}}\mathcal{L}(\mathcal{D})\approx & \frac{1}{L}\sum_{l,n=1}^{L,N}\nabla_{\boldsymbol{W}}\mathrm{log}\,p(\boldsymbol{d}_{n}|\boldsymbol{Z}^{(l)})-\sum_{i,j}\nabla_{\tilde{\boldsymbol{\pi}},\tilde{\boldsymbol{\alpha}},\tilde{\boldsymbol{\beta}}}\mathrm{KL}\big[q(z_{ij}|\tilde{\pi}_{ij})||p(z_{ij}|\pi_{ij})\big]\\
 & -\sum_{i,j}\nabla_{\tilde{\boldsymbol{\alpha}},\tilde{\boldsymbol{\beta}}}\mathrm{KL}\big[q(\pi_{ij}|\tilde{\alpha}_{ij},\tilde{\beta}_{ij})||p(\pi_{ij}|\alpha,\beta)\big]\\
 & -\varpi\sum_{i,j}\nabla_{\tilde{\boldsymbol{\pi}},\tilde{\boldsymbol{\alpha}},\tilde{\boldsymbol{\beta}}}[\mathrm{log}\,q(z_{ij}|\tilde{\pi}_{ij})+q(\pi_{ij}|\tilde{\alpha}_{ij},\tilde{\beta}_{ij})]
\end{aligned}
\end{equation}
where $\varpi$ is defined in (10). As one can note, we do not perform
Bayesian inference for the synaptic weight parameters $\boldsymbol{W}$.
Instead, we obtain point-estimates, similar to conventional DropConnect.

\subsection{Feedforward computation in DNNs with DropConnect++ layers}

Computation of the output of a trained DNN with DropConnect++ layers,
given some network input $\boldsymbol{x}_{*}$, requires that we come
up with an appropriate solution to the problem of computing the posterior
expectation of the DropConnect++ layers output, say $\boldsymbol{r}_{*}$. 

Let us consider a DropConnect++ layer with input $\boldsymbol{v}_{*}$
(corresponding to a DNN input observation $\boldsymbol{x}_{*}$);
we have
\begin{equation}
\boldsymbol{r}_{*}=\mathbb{E}_{q(\boldsymbol{Z})}[a((\boldsymbol{Z}\circ\boldsymbol{W})\boldsymbol{v}_{*})]
\end{equation}
This computation essentially consists in marginalizing out the layer
synaptic connectivity structure, by appropriately utilizing the variational
posterior distribution $q(\boldsymbol{Z})$, learned by means of BBVI,
as discussed in the previous Section. Unfortunately, this posterior
expectation cannot be computed analytically, due to the nonlinear
activation function $a(\cdot)$. 

This problem can be solved by approximating (17) via simple MC sampling:
\begin{equation}
\boldsymbol{r}_{*}\approx\frac{1}{L}\sum_{l=1}^{L}a((\boldsymbol{Z}^{(l)}\circ\boldsymbol{W})\boldsymbol{v}_{*})
\end{equation}
where the $\boldsymbol{Z}^{(l)}$ are drawn from $q(\boldsymbol{Z})$.
However, an issue such an approach suffers from is the need to retain
in memory large sample matrices $\{\boldsymbol{Z}^{(l)}\}_{l=1}^{L}$,
that may comprise millions of entries, in cases of large-scale DNNs.
To completely alleviate such computational efficiency issues, in this
work we opt for an alternative approximation that reads
\begin{equation}
\boldsymbol{r}_{*}\approx a((\tilde{\boldsymbol{\Pi}}\circ\boldsymbol{W})\boldsymbol{v}_{*})
\end{equation}
where the matrix $\tilde{\boldsymbol{\Pi}}=[\tilde{\pi}_{ij}]_{i,j}$
is obtained from the model training algorithm, described previously.
Note that such an approximation is similar to the solution adopted
by Dropout \cite{dropout}, which undoubtedly constitutes the most
popular DNN regularization technique to date. We shall examine how
this solution compares to MC sampling in the experimental section
of this work.

\begin{table*}
\begin{centering}
\caption{Predictive accuracy (\%) of the evaluated methods.}
\par\end{centering}
\centering{}%
\begin{tabular}{|c|c|c|c|c|}
\hline 
Method & CIFAR-10 & CIFAR-100 & SVHN & NORB\tabularnewline
\hline 
\hline 
No regularization & 74.47 & 41.96 & 90.53 & 90.55\tabularnewline
\hline 
Dropout & 75.70 & 46.65 & 92.14 & 92.07\tabularnewline
\hline 
DropConnect & 76.06 & 46.12 & 91.41 & 91.88\tabularnewline
\hline 
DropConnect++ & 76.54 & 47.01 & 91.99 & 93.75\tabularnewline
\hline 
\end{tabular}
\end{table*}

\begin{table*}
\begin{centering}
\caption{Computational complexity per iteration at training time ($L=1$).}
\par\end{centering}
\centering{}%
\begin{tabular}{|c|c|c|c|c|}
\hline 
\#Method & CIFAR-10 & CIFAR-100 & SVHN & NORB\tabularnewline
\hline 
\hline 
No regularization & 9s & 10s & 15s & 5s\tabularnewline
\hline 
Dropout & 9s & 10s & 15s & 5s\tabularnewline
\hline 
DropConnect & 9s & 10s & 15s & 5s\tabularnewline
\hline 
DropConnect++ & 10s & 13s & 19s & 6s\tabularnewline
\hline 
\end{tabular}
\end{table*}

\section{Experimental Evaluation}

To empirically evaluate the performance of our approach, we consider
a number of supervised learning experiments, using the CIFAR-10, CIFAR-100,
SVHN, and NORB benchmarks. In all our experiments, the used datasets
are normalized with local zero mean and unit variance; no other pre-processing
is implemented in this work\footnote{Hence, our experimental setup is \emph{not completely} \emph{identical}
to that of related works, e.g. \cite{dropconnect}; these employ more
complex pre-processing for some datasets.}. To obtain some comparative results, apart from our method we also
evaluate in our experiments DNNs with similar architecture but: (i)
application of no regularization technique; (ii) regularized via Dropout;
and (iii) regularized via DropConnect.

In all cases, we use Adagrad with minibatch size equal to 128. Adagrad's
global stepsize is chosen from the set $\{0.005,0.01,0.05\}$, based
on the network performance on the training set in the first few iterations\footnote{We have found that Adagrad allows for the best possible network regularization
by drawing just one sample per minibatch; that is, we use $L=1$ at
training time. This alleviates the training costs of both DropConnect
and DropConnect++. We train all networks for 100 epochs; we do not
apply L2 weight decay. }. The units of all the postulated DNNs comprise ReLU nonlinearities
\cite{relu}. Initialization of the network parameters is performed
via Glorot-style uniform initialization \cite{glorot}. To account
for the effects of random initialization on the observed performances,
we repeat our experiments 50 times; we report the resulting mean accuracies,
and run the Student's-t statistical significance test to examine the
statistical significance of the reported performance differences.

Prediction generation using our method is performed by employing the
efficient approximation (19). The alternative approach of relying
on MC sampling to perform feedforward computation {[}Eq. (18){]} is
evaluated in Section 4.2. In all cases, we set the prior hyperparameters
of DropConnect++ to $\alpha=\beta=1$; this is a convenient selection
which reflects that we have no preferred values for the priors $\pi_{ij}$.
The Dropout and DropConnect rates are selected on the grounds of performance
maximization, following the selection procedures reported in the related
literature. Our source codes have been developed in Python, using
the Theano\footnote{\url{http://deeplearning.net/software/theano/}}
\cite{Bastien-Theano-2012} and Lasagne\footnote{\url{https://github.com/Lasagne/Lasagne}.}
libraries. We run our experiments on an Intel Xeon 2.5GHz Quad-Core
server with 64GB RAM and an NVIDIA Tesla K40 GPU.

\subsubsection*{CIFAR-10}

The CIFAR-10 dataset consists of color images of size 32$\times$32,
that belong to 10 categories (airplanes, automobiles, birds, cats,
deers, dogs, frogs, horses, ships, trucks). We perform our experiments
using the available 50,000 training samples and 10,000 test samples.
All the evaluated methods comprise a convolutional architecture with
three layers, 32 feature maps in the first layer, 32 feature maps
in the second layer, 64 feature maps in the third layer, a $5\times5$
filter size, and a max-pooling sublayer with a pool size of 3$\times$3.
These three layers are followed by a dense layer with 64 hidden units,
regularized via Dropout, DropConnect, or DropConnect++. The resulting
performance statistics (predictive accuracy) of the evaluated methods
are depicted in the first column of Table 1. As we observe, our approach
outperforms all the considered competitors. 

\subsubsection*{CIFAR-100}

The CIFAR-100 dataset consists of 50,000 training and 10,000 testing
color images of size 32$\times$32, that belong to 100 categories.
We retain this split of the data into a training set and a test set
in the context of our experiments. The trained DNN comprises three
convolutional layers of same architecture as the ones adopted in the
CIFAR-10 experiment, that are followed by a dense layer comprising
512 hidden units. As we show in Table 1, our approach outperforms
all its competitors, yielding the best predictive performance. Note
also that the DropConnect method, which is closely related to our
approach, yields in this experiment worse results than Dropout. 

\subsubsection*{SVHN}

The Street View House Numbers (SVHN) dataset consists of 73,257 training
and 26,032 test color images of size 32x32; these depict house numbers
collected by Google Street View. We retain this split of the data
into a training set and a test set in the context of our experiments,
and adopt exactly the same DNN architecture as in the CIFAR-100 experiment.
As we show in Table 1, our method improves over the related DropConnect
method.

\subsubsection*{NORB}

The NORB (small) dataset comprises a collection of stereo images of
3D models that belong to 6 classes (animal, human, plane, truck, car,
blank). We downsample the images from 96$\times$96 to 32$\times$32,
and perform training and testing using the provided dataset split.
We train DNNs with architecture similar to the one adopted in the
context of the SVHN and CIFAR-100 datasets. As we show in Table 1,
our method outperforms all the considered competitors. 

\begin{table*}
\begin{centering}
\caption{Variation of the predictive accuracy (\%) of the MC-driven approach
(18) with the number of MC samples.}
\par\end{centering}
\centering{}%
\begin{tabular}{|c|c|c|c|c|}
\hline 
\#Samples, $L$ & CIFAR-10 & CIFAR-100 & SVHN & NORB\tabularnewline
\hline 
\hline 
1 & 74.57 & 43.28 & 91.32 & 90.04\tabularnewline
\hline 
30 & 75.95 & 46.33 & 91.70 & 90.78\tabularnewline
\hline 
50 & 76.01 & 46.33 & 91.72 & 91.04\tabularnewline
\hline 
100 & 76.01 & 46.54 & 91.78 & 91.41\tabularnewline
\hline 
500 & 76.36 & 46.94 & 91.78 & 91.58\tabularnewline
\hline 
\end{tabular}
\end{table*}

\subsection{Computational complexity}

Another significant aspect that affects the efficacy of a regularization
technique is its final computational costs, and how they compare to
the competition. To allow for investigating this aspect, in Table
2 we illustrate the time needed to complete one iteration of the training
algorithms of the evaluated networks in our implementation. As we
observe, the training algorithm of our approach imposes an 11\%-30\%
increase in the computational time per iteration, depending on the
sizes of the network and the dataset. Note though that DNN training
is an offline procedure; hence, a relatively small increase in the
required training time is reasonable, given the observed predictive
performance gains. 

On the other hand, when it comes to using a trained DNN for prediction
generation (test time), we emphasize that the computational costs
of our approach are exactly the same as in the case of Dropout. This
is, indeed, the case due to our utilization of the approximation (19),
which results in similar feedforward computations for DropConnect++
as in the case of Dropout. 

\subsection{Further investigation}

A first issue that requires deeper investigation concerns the statistical
significance of the observed performance differences. Application
of the Student's-t test on the obtained sets of performances of each
method (after 50 experiment repetitions from different random starts)
has shown that these differences are statistically significant among
all relevant pairs of methods (i.e. DropConnect++ vs. DropConnect,
DropConnect++ vs. DropOut, and DropConnect++ vs. no regularization);
only exception is the SVHN dataset, where DropConnect++ and DropOut
are shown to be of statistically comparable performance.

Further, in Table 3 we show how the predictive performance of DropConnect++
changes if we perform feedforward computation via MC sampling, as
described in Eq. (18). As we observe, using only one MC sample results
in rather poor performance; this changes fast as we increase the number
of samples. However, it appears that even with a high number of drawn
samples, the MC-driven approach (18) does not yield any performance
improvement over the approximation (19), despite imposing considerable
computational overheads.

Further, in Fig. 1(a) we illustrate predictive accuracy convergence;
for demonstration purposes, we consider the experimental case of the
CIFAR-10 benchmark. Our exhibition concerns both application of the
approximate feedforward computation rule (19), as well as resorting
to MC sampling. We observe a clear and consistent convergence pattern
in both cases.

Finally, it is interesting to get a feeling of the values that take
the inferred posterior probabilities, $\tilde{\boldsymbol{\pi}}$,
of synaptic connectivity. In Fig. 1(b), we illustrate the inferred
values of $\tilde{\boldsymbol{\pi}}$ for all the network synapses,
in the case of the CIFAR-10 experiment. As we observe, out of the
almost 300K synapses, around 50K take values less than 0.35, another
50K take values greater than 0.6, while the rest 200K take values
approximately in the interval $[0.4,0.6]$. This implies that, out
of the total 300K postulated synapses, almost half\emph{ }of them
are most likely to be omitted during inference. Most significantly,
this figure depicts that our approach \emph{infers (in a data-driven
fashion) which specific }synapses are most useful to the network (thus
yielding relatively high values of $\tilde{\pi}_{ij}$), and which
should rather be omitted. This is in contrast to existing approaches,
which merely apply a \emph{homogeneous,} \emph{random} omission/retention
rate on each layer. 

\begin{figure*}
\noindent \begin{centering}
\subfloat[]{\includegraphics[width=4.5cm,height=2.6cm]{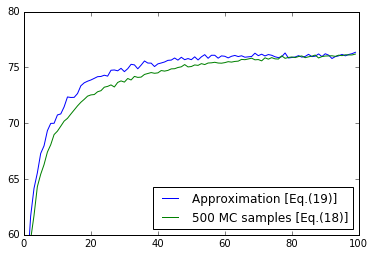}}\hfill{}\subfloat[]{\includegraphics[width=4.5cm,height=2.6cm]{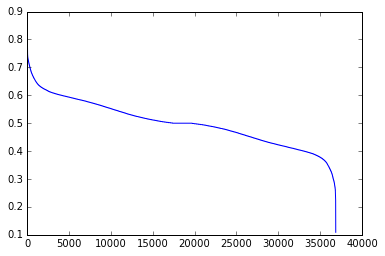}}
\par\end{centering}
\noindent \centering{}\caption{(a) Accuracy convergence. (b) Inferred posterior probabilities, $\tilde{\boldsymbol{\pi}}$.}
\end{figure*}

\section{Conclusions}

In this paper, we examined whether there is a feasible way of performing
DNN regularization by marginalizing over network synaptic connectivity
in a Bayesian manner. Specifically, we sought to derive an appropriate
posterior distribution over the network synaptic connectivity, inferred
from the data. To this end, we imposed a sparsity-inducing prior over
the network synaptic weights, where the sparsity is induced by a set
of Bernoulli-distributed binary variables. Further, we imposed appropriate
Beta (hyper-)priors over the parameters of the postulated Bernoulli-distributed
binary variables. Under this hierarchical Bayesian construction, we
obtained appropriate posteriors over the postulated binary variables,
which indicate which synaptic connections are retained and which or
dropped during inference. This was effected in an efficient and elegant
fashion, by resorting to BBVI. We performed an extensive experimental
evaluation, using several benchmark datasets. In most cases,
our approach yielded a statistically significant
performance improvement, for competitive computational costs.

\subsubsection*{Acknowledgment}

We gratefully acknowledge the support of NVIDIA Corporation with the
donation of one Tesla K40 GPU used for this research.

\section*{Appendix}

\begin{equation}
\begin{aligned}\mathrm{KL}[q(z_{ij}|\tilde{\pi}_{ij})||p(z_{ij}|\pi_{ij})\big] & =\tilde{\pi}_{ij}\mathrm{log}\tilde{\pi}_{ij}+(1-\tilde{\pi}_{ij})\mathrm{log}(1-\tilde{\pi}_{ij})\\
 & -\tilde{\pi}_{ij}\mathbb{E}_{q(\pi_{ij})}[\mathrm{log}\pi_{ij}]-(1-\tilde{\pi}_{ij})\mathbb{E}_{q(\pi_{ij})}[\mathrm{log}(1-\pi_{ij})]
\end{aligned}
\end{equation}
\begin{equation}
\begin{aligned}\mathrm{KL}\big[q(\pi_{ij}|\tilde{\alpha}_{ij},\tilde{\beta}_{ij})||p(\pi_{ij}|\alpha,\beta)\big] & =\mathrm{log}\Gamma(\tilde{\alpha}_{ij}+\tilde{\beta}_{ij})-\mathrm{log}\Gamma(\tilde{\alpha}_{ij})-\mathrm{log}\Gamma(\tilde{\beta}_{ij})\\
+(\tilde{\alpha}_{ij}-\alpha) & \mathbb{E}_{q(\pi_{ij})}[\mathrm{log}\pi_{ij}]+(\tilde{\beta}_{ij}-\beta)\mathbb{E}_{q(\pi_{ij})}[\mathrm{log}(1-\pi_{ij})]
\end{aligned}
\end{equation}
 where:
\begin{equation}
\mathbb{E}_{q(\pi_{ij})}[\mathrm{log}\pi_{ij}]=\psi(\tilde{\alpha}_{ij})-\psi(\tilde{\alpha}_{ij}+\tilde{\beta}_{ij})
\end{equation}
\begin{equation}
\mathbb{E}_{q(\pi_{ij})}[\mathrm{log}(1-\pi_{ij})]=\psi(\tilde{\beta}_{ij})-\psi(\tilde{\alpha}_{ij}+\tilde{\beta}_{ij})
\end{equation}
 $\Gamma(\cdot)$ is the Gamma function, and $\psi(\cdot)$ is the
Digamma function.

\bibliographystyle{splncs03}
\bibliography{dc++}

\begin{thebibliography}{10}
\providecommand{\url}[1]{\texttt{#1}}
\providecommand{\urlprefix}{URL }

\bibitem{understandDropout}
Baldi, P., Sadowski, P.: Understanding dropout. In: Proc. NIPS (2013)

\bibitem{Bastien-Theano-2012}
Bastien, F., Lamblin, P., Pascanu, R., Bergstra, J., Goodfellow, I.J.,
  Bergeron, A., Bouchard, N., Bengio, Y.: Theano: new features and speed
  improvements. Deep Learning and Unsupervised Feature Learning NIPS 2012
  Workshop (2012)

\bibitem{dae}
Bengio, Y., Yao, L., Alain, G., Vincent, P.: Generalized denoising autoencoders
  as generative models. In: Proc. NIPS. pp. 899-- 907 (2013)

\bibitem{adagrad}
Duchi, J., Hazan, E., Singer, Y.: Adaptive subgradient methods for online
  learning and stochastic optimization. J. Machine Learning Research  12,
  2121-- 2159 (2010)

\bibitem{bayes-dropout}
Gal, Y., Ghahramani, Z.: Dropout as a {Bayesian} approximation: Insights and
  applications. In: Deep Learning Workshop, ICML (2015)

\bibitem{glorot}
Glorot, X., Bengio, Y.: Understanding the difficulty of training deep
  feedforward neural networks. In: Proc. AISTATS (2010)

\bibitem{glynn}
Glynn, P.W.: Likelihood ratio gradient estimation for stochastic systems.
  Communications of the ACM  33(10),  75--84 (1990)

\bibitem{jaakola}
Jaakkola, T., Jordan, M.: Bayesian parameter estimation via variational
  methods. Statistics and Computing  10,  25--37 (2000)

\bibitem{adam}
Kingma, D., Ba, J.: Adam: A method for stochastic optimization. In: Proc. ICLR
  (2015)

\bibitem{lecun}
LeCun, Y., Bengio, Y., Hinton, G.: Deep learning. Nature  512,  436--444 (2015)

\bibitem{relu}
Nair, V., Hinton, G.: Rectified linear units improve restricted {Boltzmann}
  machines. In: Proc. ICML (2010)

\bibitem{bbvi}
Ranganath, R., Gerrish, S., Blei, D.M.: Black box variational inference. In:
  Proc. AISTATS (2014)

\bibitem{dropout}
Srivastava, N., Hinton, G.E., Krizhevsky, A., Sutskever, I., Salakhutdinov,
  R.R.: Dropout: A simple way to prevent neural networks from overfitting. J.
  Machine Learning Research  15(6),  1929--1958 (June 2014)

\bibitem{dropconnect}
Wan, L., Zeiler, M., Zhang, S., LeCun, Y., Fergus, R.: Regularization of neural
  networks using {DropConnect}. In: Proc. ICML (2013)

\bibitem{reinforce}
Williams, R.J.: Simple statistical gradient-following algorithms for
  connectionist reinforcement learning. Machine Learning  8(3-4),  229--256
  (1992)

\end{thebibliography}

\end{document}